\documentclass[sigconf]{acmart}

\usepackage{multicol}
\usepackage{multirow}
\usepackage{balance}

\AtBeginDocument{%
  \providecommand\BibTeX{{%
    \normalfont B\kern-0.5em{\scshape i\kern-0.25em b}\kern-0.8em\TeX}}}

\copyrightyear{2024}
\acmYear{2024}
\setcopyright{acmcopyright}

\acmConference[WSDM '24] {Proceedings of the 17th ACM International Conference on Web Search and Data Mining}{March 4--8, 2024}{Merida, Mexico.}
\acmBooktitle{Proceedings of the 17th ACM International Conference on Web Search and Data Mining (WSDM '24), March 4--8, 2024, Merida, Mexico}
\acmPrice{}
\acmISBN{979-8-4007-0371-3/24/03}
\acmDOI{10.1145/3616855.3635849}
\begin{document}

\title{IDoFew: Intermediate Training Using Dual-Clustering in Language Models for Few Labels Text Classification}


\author{$\text{Abdullah Alsuhaibani}^{1,5}$, $\text{Hamad Zogan}^{1,2}$, $\text{Imran Razzak}^{3}$, $\text{Shoaib Jameel}^{4}$, $\text{Guandong Xu}^{1}$}
\affiliation{
  \institution{${}^{1}$University of Technology Sydney \country{Australia}\\${}^{2}$College of Computer Science and Information Technology \country{Saudi Arabia} \\ ${}^{3}$University of New South Wales \country{Australia} \\ ${}^{4}$University of Southampton \country{United Kingdom} \\ ${}^{5}$Islamic University of Madinah \country{Saudi Arabia}}}
  \email{abdullah.alsuhaibani@student.uts.edu.au, hamad.a.zogan@student.uts.edu.au, imran.razzak@unsw.edu.au}
  \email{M.S.Jameel@southampton.ac.uk, guandong.xu@uts.edu.au}

  





\begin{abstract}
Language models such as Bidirectional Encoder Representations from Transformers (BERT) have been very effective in various Natural Language Processing (NLP) and text mining tasks including text classification. However, some tasks still pose challenges for these models, that include text classification with limited labels. This can result in a cold-start problem. Although some approaches have attempted to address this problem through single-stage clustering as an intermediate training step coupled with a pre-trained language model, which generates pseudo-labels to improve classification, these methods are often error-prone due to the limitations of the clustering algorithms. To overcome this, we have developed a novel two-stage intermediate clustering with subsequent fine-tuning that models the pseudo-labels reliably, resulting in reduced prediction errors. The key novelty in our model, IDoFew, is that the two-stage clustering coupled with two different clustering algorithms helps exploit the advantages of the complementary algorithms that reduce the errors in generating reliable pseudo-labels for fine-tuning. Our approach has shown significant improvements compared to strong comparative models.
\end{abstract}





\maketitle

\section{Introduction}
Collecting a large amount of data with annotations is a challenging task. The motivation is that if we have more data we can reliably train the machine learning model parameters so that it could perform as expected in some downstream applications. The problem is even more critical in applications where the performance that the automated system must give should remain faithful, e.g., medical applications. While this is an ideal setting, it is often difficult and not very cheap to obtain annotated instances to train a machine learning model. Over the number of years, researchers have developed techniques to address the limitations, e.g., few-shot learning, unsupervised learning and others, and in recent years, there has been a surge in developing techniques based on the self-supervised learning paradigm. By exploiting the self-supervised approaches, researchers developed a number of pre-trained language models (PTMs) that are usually first trained in a domain-independent dataset and later fine-tuned in some domain-specific target tasks \cite{devlin-etal-2019-bert}. In the fine-tuning stage, we usually rely upon high-quality annotated data to reliably fine-tune the language model. However, the fine-tuned model still generalizes poorly to out-of-distribution input data created with few and harmless perturbations \cite{ribeiro2020beyond}.
 
There is another line of research called few-shot text classification. Few labels and few-shot learning paradigms lie under the umbrella of machine learning, however, there are some key differences. In few-shot learning, the task of learning is to classify new data when we only have a few labelled samples per class. We might want to learn to classify a cat with only a few images of a cat. Few-label learning refers to the task of learning to classify new data when we have a few labels overall. For example, we might be able to learn to classify different types of diseases with only a few hundred labelled medical documents. The key difference thus is that few-shot learning is about learning from small datasets with many classes, while few-labels learning is about learning from small datasets with few classes.

In problems where there are only a few labels in a limited number of classes, language models may not perform optimally. To address this issue, researchers have previously developed intermediate training using unsupervised learning methods. For example, in a recent study \cite{shnarch-etal-2022-cluster}, a new type of intermediate task was developed that aligns naturally with text classification. This technique exploits text clustering for inter-training the language model and generates pseudo-labels, also known as weak labels. The clustering algorithm partitions the data into different salient features in the corpus, which leads to representations that are well-suited for the target text classification task. Each cluster then acts as a label to the language model, and the task is to predict the cluster label during inter-training. This technique is influenced by the computer vision community, where clustering has been commonly used to obtain pseudo-labels. However, the quality of the pseudo-labels is largely dependent on the clustering algorithm used, and if the labels are of low quality, the fine-tuning classification model may underperform.

We have developed a new computational model, IDoFew, to address a particular issue. Our model uses a two-stage clustering approach. In the first stage, we consider that the cluster labels may not be perfect, which leads to a less reliable fine-tuned language model. We use the full dataset to perform this clustering. However, unlike past works, we introduce a second stage of clustering that acts as a ``label correction engine'' using only a small subset of randomly collected instances that help correct the misprediction errors of the first stage of clustering. Essentially, we are taking advantage of the strengths of two different clustering algorithms, resulting in a more accurate final text classification model that is suitable for a few labels. Such dual-clustering methodologies have been developed in the past with success \cite{matuszewski2002double}. We have tested our model on different benchmark text collections that include both long and short texts and varying numbers of classes. Our experiment results show that our model achieves state-of-the-art results.

Our primary contribution is the creation of a new model that utilizes a two-stage clustering approach to enhance text classification in situations where there are few labels available. This innovative inter-training technique enables a language model to adapt effectively to the challenge of text classification with limited labels. Furthermore, our model is specifically designed to align with the objective of text classification, and it not only overcomes the shortcomings of previous methods but can also be easily implemented in real-world scenarios.

\section{Related Work}
In this section, we summarise the closely related and relevant models. Besides that, we also mention how our approach is different from other existing methods.

Text mining, Feldman and Dagan \cite{Feldman1995KnowledgeDI}, applies models for analyzing the text in different ways, e.g., topic detection. Bag of words (BoW) \cite{Bloehdorn2004BoostingFT} is still a popular model to represent text for input to a machine learning model and remains relevant to this day \cite{xu-etal-2015-short} in clustering. Text clustering is an unsupervised approach to automatically discover the salient feature present in data. In text classification, the goal is to group the text into a number of clusters as similar representation is brought together in one cluster. In \cite{singhania2022predicting}, the authors studied the effects of data representation where they used a BERT encoder to cluster the text.

Another important paradigm in text mining is classification, where recently, PTMs have played a major role \cite{EinDor2020CorpusWA}. In fact, PTMs have been popular even in cases when labelled data is limited and the training set is small \cite{chau-etal-2020-parsing}. Instead of using the transformer model, it is suggested that the transformer structure itself be enhanced \cite{wang-etal-2022-clusterformer}. In \cite{petroni-etal-2019-language} the authors recommended giving hundreds of labels of examples using prompting for fine-tuning language models \cite{devlin-etal-2019-bert}.


The primary goal of semi-supervised learning is to make trained models effectively improve by using an enormous amount of unlabeled data with less labelled data \cite{chapelle2009semi}. One type of semi-supervised learning is pseudo-labeling \cite{arazo2020pseudo} which generates labels from unlabeled data and is used for model training. It can excel when there is a  scarcity of labelled data \cite{shnarch-etal-2022-cluster}. In practice, NLP adopts a semi-supervised learning approach for different tasks, namely machine translation \cite{edunov-etal-2018-understanding} and text generation \cite{gupta-etal-2020-semi}. For text classification, semi-supervised learning has attracted attention recently. In \cite{xu-etal-2022-progressive}, the authors use a combination of semi-supervised learning and PTMs to inherit topic matching by updating the semantic representation and the classifier of the matching and K-way. To address the poor performance due to margin bias in semi-supervised learning in text classification, \cite{li-etal-2021-semi-supervised} proposed a Semi-Supervised Text Classification with Balanced Deep representation Distributions (S\textsuperscript{2}TC-BDD) which measures the difference for labelled and unlabelled texts. With the improvement in PLMs, self-training \cite{ zhou-etal-2021-self-supervised} is attracting increasing attention in NLP. 

Recently, there has been a surge in zero/few-shot learning for many text-related tasks such as relation classification \cite{soares2019matching}, and named entity recognition \cite{huang2021few}. Text classification under a few-shot setting performs poorly when traditional methods are used. However, various methods have been developed to boost performance, such as attention mechanisms, meta-learning and prompt learning. Working on sentence embedding level which causes some bias to particular classes, it is proposed \cite{Gao_Han_Liu_Sun_2019} to use attention to concentrate on vital dimensions. In \cite{Deng_Zhang_Sun_Chen_Chen_2020}, the authors suggested combining meta-learning with an unsupervised language model to address the adaption of unnoticeable classes or when data is insufficient. In \cite{https://doi.org/10.48550/arxiv.2109.04707}, the authors proposed KGML where the gap between training and testing falls in meta-learning as well as a few examples of text classification. Recently, prompt-based approaches have improved few-shot learning for large PLMs in relation to text classification. In \cite{wang-etal-2021-transprompt}, the authors proposed the TransPrompt framework which is to fetch transferable knowledge among the tasks. In addition, a neural network is a part of the development for such a problem. In \cite{sun-etal-2019-hierarchical}, the authors proposed hierarchical attention prototypical networks (HAPN) for few-shot text classification which is designed to enhance the expressive ability of the semantics so it can increase or diminish the importance of words, features and instances. Schick and Sch\"{u}tze applied self-training iteratively to incorporate few-shot learning and multiple-generation models were trained on pseudo-labelled data by previous-generation ensembles \cite{schick2020exploiting}. The amount of data required to train the model was comparatively large. In other work, Du et al., focused on retrieving more task-related data (unlabelled) from the open domain corpus, however, it is dependent on a quality sentence encoder \cite{du2020self}.

There are works where there is an intermediary step to fine-tune models into targeted data points by training using the source data first \cite{choshen2022start}. In \cite{shnarch-etal-2018-will,yu-etal-2021-fine}, the authors developed weak label data that was used during training. After the source tasks were followed by fine-tuning on target tasks, it was noticed that after inter-training some linguistic knowledge variations were modelled \cite{pruksachatkun-etal-2020-intermediate}. Recently, \cite{shnarch-etal-2022-cluster} investigated the intermediate tasks along with the BERT model. However, there were certain issues with the model such as clustering was unreliable and the model was inefficient. We address these issues in this paper.

In this work, we present a novel model Intermediate Dual-Clustering for Few Labels in text classification, IDoFew, which improves text classification when there are only a few labels. To the best of our knowledge, we are the first to use inherited knowledge with small amounts of data to be trained within clusters with PTM for text classification tasks when there are a limited number of labels.    


\begin{figure}
  \includegraphics[scale=0.35]{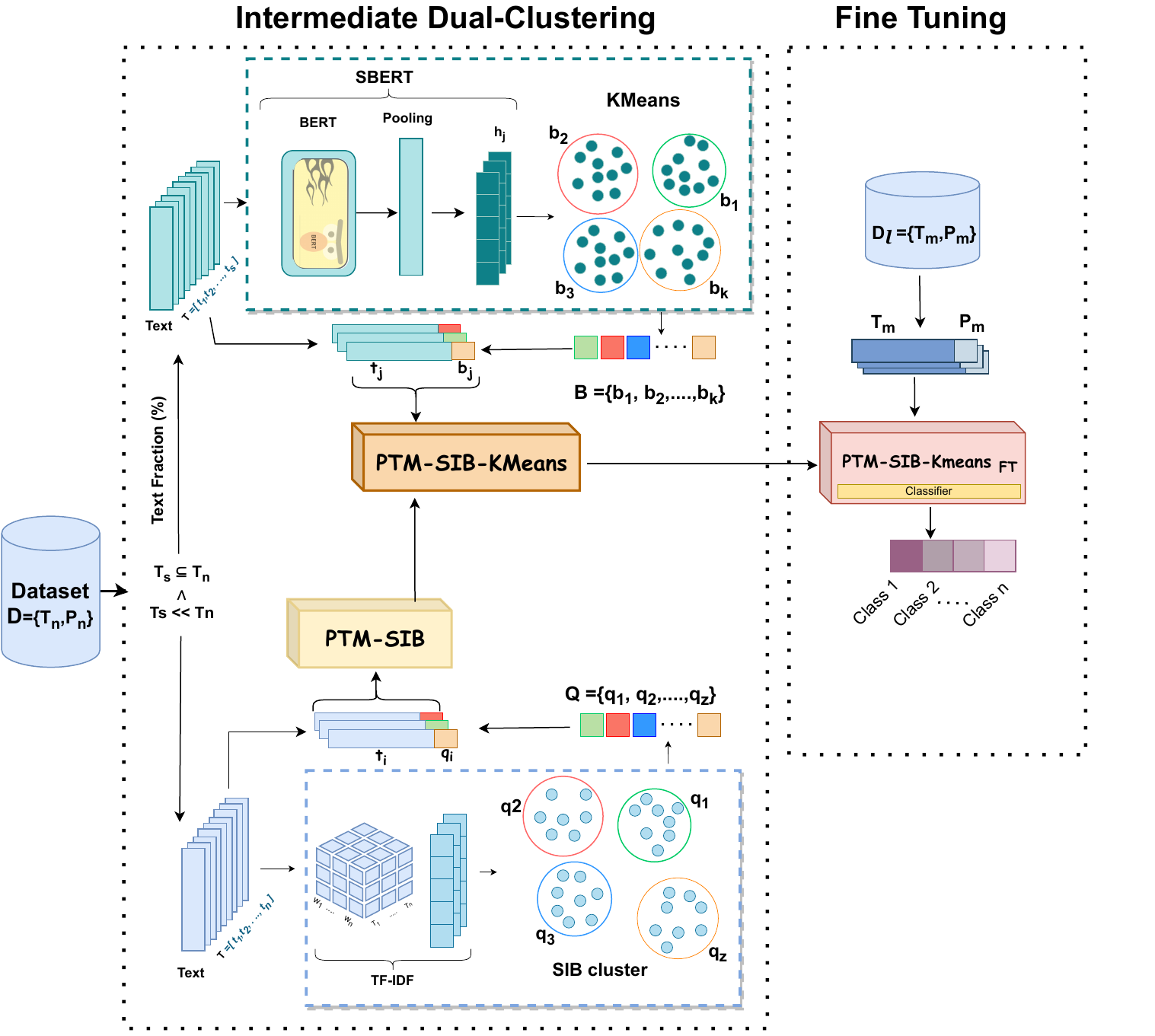 }
   \caption{l Intermediate Dual-Clustering
for Few Labels in text classification (IDoFew). PTM indicate to pre-trained models -- BERT, RoBERTa, and DistilBERT.  Dash-dots produce the pseudo labels for each model.}
  \label{fig:model_ICFT}
\end{figure}

\section{IDoFew: Our Novel Model}
\noindent \emph{\textbf{Motivation:}} In this section, we describe the mathematical details of our model. The key motivation of our model is to improve the text classification performance on a few label problems. To this end, we develop a novel model that comprises two stages of clustering. In the first stage, we use the full dataset and cluster the instances to obtain pseudo-labels. Our assumption is that, like previous work, these pseudo-labels might not be ideal to give a faithful quantitative performance when inter-training with the PTM. To correct the possible mistakes in the first layer of clustering, we develop a second layer of clustering approach using only a small subset of the samples that are randomly chosen from the dataset. This subset of samples is then clustered with another simple clustering approach to reduce the overall computational burden and improve the performance of the model. The advantage of adopting a lightweight second clustering approach is that it helps correct any mistakes made by the first stage of clustering. As a result, we see an overall gain in the quantitative performance. Figure~\ref{fig:model_ICFT} illustrates our novel model.

\noindent \emph{\textbf{Notations:}} Our framework comprises input data $D=\{T_n\}$ where $T_n$ and $P_n$ are texts and corresponding labels, respectively. The text is represented by $T=[t_1,t_2,....,t_n]$ where $n$ is a text dimension of the input dataset. Assume $D_l=\{T_m,P_m\}$ are the labelled samples of data $T_m$ and $P_m$ is text and the corresponding label, respectively such that $D_l << D$. Our aim is to inter-train the PTM so that it is suitable for a few-label classification model. We denote the dataset as $D$ that contains $T_n$ instances, each comprising of \(N\) sentences $\{t_i\}^\mathbf{N}_{i=1}$.

\noindent \emph{\textbf{PTM-SIB}}: In the first stage of clustering, we have a clustering model that needs input feature representations. The goal is to learn the pseudo-labels automatically. A common approach in the literature is to use the term-frequency and inverse-document frequency (TF-IDF). TF-IDF is a numerical statistic metric that reflects a word's combined inter and intra-document importance in a corpus. Consider the texts in the BoW paradigm, where each text is represented as a series of words without regard to their order. According to BoW, a word with a frequency of 10 occurrences in a text is more significant than a word with a frequency of 1, but not in proportion to the frequency. The log term frequency $ltf$ used in this context is defined by:

\begin{equation}
\label{equ:tf}
ltf_{(r,t_i)} = 1+log(1+tf_{(r,t_i)})
\end{equation}

\noindent where $tf_{(r,t_i)}$ represents the occurrence number of the term $r$ in a text $t_i$. The IDF uses the frequency of the term in the whole collection to weigh the term's significance in light of the inverse effect. The log IDF is an indicator of a term's informativeness and is defined as follows:

\begin{equation}
\label{equ:idf}
idf_{r}=log \frac{N}{df_{r}}
\end{equation}

\noindent where $N$ represents the full text, and $df_{r}$ the number of texts containing the term $r$. To calculate TF-IDF, TF and IDF are combined as follows: 

\begin{equation}
\label{equ:tf-idf}
tfidf_{(r,t_i)}= ltf_{(r,t_i)} * idf_{r}
\end{equation}

Subsequently, we utilize unsupervised learning and generate labels (pseudo-labels) to create new knowledge based on our main assumption. Despite the abundance of clustering algorithms utilized in NLP tasks \cite{KIM2020113288}, it is crucial to select one that can effectively handle text data in a practical model. Therefore, we use the sequential Information Bottleneck (SIB) cluster \cite{10.1145/564376.564401} as an unsupervised algorithm. SIB covers a local maximum of document information with respect to time and memory space and works effectively with datasets of different sizes. Starting with a random partition of a cluster, at each step SIB draws samples of $X_s$ randomly deals with each as a singleton cluster and joins it to produce a new partition. To select a new cluster for a sample $X_i$ the Jensen-Shannon divergence distance function is used. Each step can be improved or it can remain unchanged for the score function.

The core aim of the SIB cluster is to create $z$ clusters that perform pseudo-labeling. The cluster is represented by $Q= [q_1 ,...,q_z]$. Hence, each cluster in $Q$ represents a new label for each text. Therefore, the pseudo label $q_i$ and the text $t_i$ are used as input to train PTM. During training, the pre-trained model parameters are updated when trained on the pseudo-labels, and the updated model is called PTM-SIB and is used for the next stage.

\noindent \emph{\textbf{PTM-SIB-KMeans:}} In this stage, the primary aim, following the acquisition of knowledge from the prior box model, particularly cluster methodology, is to transfer that knowledge through a process of testing and refinement. This stage is similar to the structure of PTM-SIB. However, unlike PTM-SIB, we only use a small fraction of the dataset\footnote{Fraction is based on percentage used (e.g. 5\%, 10\%, or 20\%)}, which helps to overcome the data imbalance problem and avoids smaller clusters being wrongly merged and larger clusters being wrongly split. Table \ref{tbl:bert-sib-comp} validates the utilization of a fraction of the dataset for this stage. 

For correction mechanism purposes, we replace word embedding with sentence embedding. To this end, we exploit the Sentence-BERT (SBERT) \cite{reimers-gurevych-2019-sentence} model. Their embedding model is based on the BERT architecture. 

SBERT uses the output for pooling operation by computing mean-strategy to get fixed-size sentence vectors. We represent our dataset $D$ that contains $T=[t_{1}, ...., t_{s}]$, where $T_s$ denotes a set of random text using the fraction method. We feed each sentence $t_i$ into SBERT and compute the token-level hidden representations  $h_{i,s} \in \mathbb R^\mathbf {len(t_i) \times d}$:

\begin{equation}
\label{equ1}
[h_0;h_1;.....;h_S;....;h_l] =SBERT(t_i)
\end{equation}

\noindent where $len(t_i)$ is the length of the tokenized sentence, $d$ is the size of SBERT's hidden representations, and $l$ is the number of hidden layers in SBERT. The embedded summary for the set of random text $T_s$ can be represented as $=[H_{1},H_{2},H_{3}, ...., H_{S}]$.

The output of embedding is passed to the cluster, using the KMeans algorithm that generates pseudo-labels where each cluster represents a pseudo-label. It is shown that KMeans is an effective technique for clustering features \cite{REN20221}. Moreover, KMeans performs better when the number of examples is relatively small. Assume $B=\{b_1,...,b_k\}$, is a set of $k$ clusters that partition $H$ where $k$ is a positive integer greater than one, and the clusters $b_j$ represent a new label for text $t_j$. The mean $u_j$ of data-points in $b_j$ is defined as:
 
\begin{equation} u_{j}=\frac{1}{\vert b_{j}\vert }\sum\limits_{H\in b_{j}}H  \end{equation}
 
\noindent where $|b_j|$ is the cardinality in $b_j$. 

The goal of the KMeans algorithm is to determine the minimum sum of squared errors over all \(K\) clusters and it can be calculated as follows:
 
\begin{equation} \arg_{B}\min\sum\limits_{j=1}^{K}\sum\limits_{H{\in b}_{j}}\Vert H-u_{k}\Vert ^{2}
\end{equation} 

A second clustering algorithm is used instead of using a clustering algorithm for more than one epoch which is time-consuming \cite{shnarch-etal-2022-cluster}. In addition, the portion of fractions works as corrections as explained in the analysis section. In light of this, our model demonstrates better performance. We use the $B_s$ and $T_s$, which represent text and a generated label respectively to train the PTM-SIB model. As a result, when trained on the label obtained from the second cluster, very few changes are applied to the PTM-SIB configuration, so we called the newly produced model, PTM-SIB-KMeans and it is used for the next stage.

\noindent \emph{\textbf{Fine-Tune with few labels:}} Fine-tuning the model is the last stage in the proposed model. As a word embedding, default PTM embedding is used. Let $D_l=\{T_m,P_m\}$, which denotes a small portion of labelled input data, i.e. $D_l << D$, where $T_m$ is the text and $P_m$ is the true labels. We represent $m$ texts as $T = [t_1, t_2,..., t_m]$, and the corresponding label as $P= [p_1,p_2,...,p_m]$. PTMs are characterised by an encoder that creates a hidden state contextualized vector representation for each token in $t_i$. The next step is to classify the text using our pre-trained PTM-SIB-KMeans. In classifying the input text, we thus added a fully connected neural network to the PTM-SIB-KMeans model's output. This classifier is only used with the special [CLS] token's final hidden state vector. Since we are using a few labels in this stage we call it PTM-SIB-KMeans$_{FT}$.  

\section{Experiments and Results}
This section briefly describes the datasets used, the baseline methods, and the experiment setup.

 \begin{table}
 
 \centering
  \caption{Datasets and statistical information}
  \label{tab:data}
  \begin{tabular}{lcc} 
    \hline    \hline
    \textbf{Datasets} & \textbf{Num of Classes}&\textbf{Size}\\
    \midrule    \\ 
    Subjectivity & 2 & 8,000\\
    Polarity & 2  &    8,500\\
    SMS Spam & 2 &     5,000\\
    ISEAR & 7 &      6,000\\
    AG News & 4 &      17,143\\
    Yahoo Answers & 10 & 17,143\\
    DBpedia& 14& 17,143\\
  \hline
    \hline
\end{tabular}
\end{table}

\begin{table*}
\centering
\caption{Our model, IDoFew, performance accuracy(\%) compared to baselines models and BERT$_{IT:CLUSTER}$. Note that $_{FT}$ indicates the fraction of text used in the second stage and fine-tuning is on 64 samples for all models.}
\label{tab:result}
\begin{tabular}{lllcllll}
\hline \hline
Models / Datasets                                                                                     & SMS Spam       & Polarity       & Subjectivity   & Yahoo  & ISEAR          & AG News        & DBpedia \\ \hline 
\multicolumn{8}{c}{\textbf{Baseline }} \\

 BERT                                                                                                     & 86.20          & 66.26          & 92.7           & 15.77          & 17.31          & 51.00          & 31.54   \\
RoBERTa                                                                                                & 84.12          & 61.10          & 72.1           & 11.24          & 21.57          & 58.56          & 20.95   \\
DistilBERT                                                                                                & 83.29          & 66.72          & 88.6           & 17.21          & 22.9           & 64.48          & 23.09   \\ 

\hline
\multicolumn{8}{c}{\textbf{Baseline Zero label }} \\

 BERT                                                                                                     & 80.66          & 48.54          & 50.80           & 10.45          & 15.31          & 24.91          & 7.51   \\
RoBERTa                                                                                                & 81.84          & 48.45          & 50.74           & 10.07          & 12.64          & 22.11          & 8.58   \\
DistilBERT                                                                                                & 85.29          & 51.54          & 49.32           & 9.42          & 13.44           & 35.13          & 10.31   \\ 

\hline
\multicolumn{8}{c}{\textbf{State-of-the-art Few labels frameworks }} \\
 BERT{$_{IT:CLUSTER}$}                                                                        & 98.20          & 67.02          & 92.84          & 47.46          & 30.09          & 82.90          & 68.00   \\
 BERT{$_{IT:MLM}$}                                                                        & 89.53          & 66.25          & 93.01          & 20.78          & 22.89          & 77.63          & 55.10   \\
 BERT{$_{IT:MLM+CLUSTER}$}                                                                       & 98.73          & 74.82          & 93.00          & 49.52          & 32.17          & 83.74          & 74.33   \\
\hline
\multicolumn{8}{c}{\textbf{{ Intermediate Dual-Clustering for Few Labels (IDoFew)}}} \\
\multicolumn{1}{l}{Be-SIB-KMeans$_{FT}$ } & \textbf{98.74} & 69.82 & \textbf{93.80} & 52.54 & \textbf{32.88} & \textbf{84.64} & \textbf{77.41}   \\

RoB-SIB-KMeans$_{FT}$ & 98.20 & \textbf{75.82} & 92.00 & \textbf{54.54} & 32.62 & 82.64 & 74.42 \\

DisB-SIB-KMeans$_{FT}$ & 96.95 & 63.26 & 91.60 & 40.83 & 27.29 & 78.86 & 73.02 \\

\hline \hline
\end{tabular}
\end{table*}

\subsection{Datasets}
We have used seven datasets in our work to evaluate the effectiveness of our technique in comparison with other state-of-the-art models. These datasets are: \textbf{SMS Spam} \cite{Almeida2011SpamFiltering} which is a dataset that categorizes a message into spam or ham (not spam). \textbf{Polarity} \cite{pang-lee-2005-seeing} is a dataset containing movie reviews that are either positive or negative. \textbf{Subjectivity} \cite{Pang+Lee:04a} is a dataset containing movie reviews which are either subjective or objective. \textbf{AG News}  \cite{NIPS2015_250cf8b5} is a dataset which groups news articles into four classes. \textbf{Yahoo Answers} \cite{NIPS2015_250cf8b5} is a question-answering dataset divided into ten topic classes. \textbf{DBpedia} \cite{NIPS2015_250cf8b5} is a dataset of Wikipedia articles that are grouped by entity types. \textbf{ISEAR}\cite{395649e6ec0148c58662364553c56e48} is a dataset of student response reports that are labelled based on seven major emotions. Table \ref{tab:data} presents more information about the datasets.

\begin{table}
 \centering
  \caption{Comparing to the state-of-the-art few shot model on Subjectivity and AG News datasets.}
  \label{tab:data_subj}
  \begin{tabular}{lcc}
    \hline    \hline
    \textbf{Model} & \textbf{Subjectivity}& \textbf{AG New}\\
    \midrule    \\ 
    LM-BFF & 89.64 & 81.08\\
    PET & 91.88    & 81.41 \\
    SFLM & 93.25    &  83.26\\ 
    \hline
    Proposed (Be-SIB-KMeans) & \textbf{93.80 }&  \textbf{84.64}\\
  \hline
    \hline
\end{tabular}
\end{table}

All the datasets are split into 80\% for the train set and 20\% for the test set. Since some datasets are relatively large, which does not fit our model motivation, a random sample is selected from each dataset \cite{shnarch-etal-2022-cluster}. We preprocess the text by removing English stop words and punctuation that are present in the text. Additionally, we convert the text to lowercase as required by the clustering algorithms. The annotations that come with the instances are utilized during the fine-tuning stage of the PTM after the cluster inter-training is complete.

\subsection{Comparative Models}
The pre-trained models BERT, RoBERTa and DistilBERT \cite{devlin-etal-2019-bert, sanh2020distilbert, liu2019roberta} are used as the baseline models to compare with our model. We have used the base-uncased models for BERT and DistilBERT, and the base model for RoBERTa. BERT and RoBERTa contain 12 transformer block layers and DistilBERT has six layers, with a head size of 786 with 12 self-attention heads and 110M, 125M, and 66M parameters, respectively. As an optimizer, we use the Adam algorithm \cite{Kingma2015AdamAM} and the learning rate = $3e-5$  \cite{devlin-etal-2019-bert}. A batch of size 64 and 10 epochs are used. Exact setups are used for all the models to ensure a fair comparison. As the main state of the art in intermediate training, we compare our model to \cite{shnarch-etal-2022-cluster} where they explore three models: BERT{$_{IT:CLUSTER}$}, BERT{$_{IT:MLM}$}, and BERT{$_{IT:MLM+CLUSTER}$}. We implement our version based on hyper-parameters recommended in their work. Besides a few label settings, we have also compared with state-of-the-art few-shot learning settings such as \cite{schick-schutze-2021-exploiting}, LM-BFF \cite{gao-etal-2021-making}, and SFLM \cite{chen-etal-2021-revisiting}. 



\subsection{IDoFew Experiment Settings}
\textbf{Word and Sentence Embedding:}  The top 10,000 vocabulary words generated by TF-IDF are employed as word embeddings for the SIB cluster algorithm within the PTM-SIB stage. In the subsequent stage, PTM-SIB-KMeans, which integrates with KMeans clustering, sentence embeddings are crafted using SBERT, specifically all-MiniLM-L6-v2. SBERT is a sentence transformer model adept at capturing semantic information. It is trained on 1 billion sentence pairs and transforms sentences into 384-dimensional dense vectors. Sentences exceeding 256 words are truncated to fit this size. Three pre-trained language models served as base models: BERT, RoBERTa, and DistilBERT. These models are among the most extensively utilized pre-trained models in natural language processing and have demonstrated remarkable performance across various NLP tasks.

\textbf{Unsupervised  Algorithms:} Two distinct clustering algorithms were employed for this study. Our novel model utilized SIB with a maximum iteration of 15, while KMeans employed 300 iterations as the default setup. The number of clusters is a critical parameter in this work and was set to $C$= 25 for both algorithms, with the exception of DBpedia where $C$=25 was used.

\textbf{Text Input Size and Few Labels:} The text input size is independently managed within the proposed structure for the PTM-SIB, PTM-SIB-KMeans, and Fine-Tune stages. Due to varying training set sizes, a portion of the text is passed to unsupervised pre-trained models instead of fixed-size inputs. The PTM-SIB stage operates on the entire text (100\%) denoted by $T$. The text $T_s$ in PTM-SIB-KMeans is randomly selected from the dataset's total input, ranging from 5\% to 20\%. This approach addresses the data imbalance issue and prevents smaller clusters from being erroneously merged or larger clusters from being mistakenly split. A more detailed analysis is presented in the ablation study.

During the final stage, fine-tuning is exposed to the actual label $P_m$ alongside text $T_m$, as depicted in Figure \ref{fig:model_ICFT}. This work employs 64 samples as a fixed parameter. Note that PTM-SIB-KMeans$_{FT}$ denotes the final model, where $_{FT}$ represents the fraction of text (5\%) employed in the second intermediate training. We utilize the accuracy metric to assess all models. Our experiment's findings demonstrate that our model outperforms both the state-of-the-art models and the baseline models.



\subsection{Results}

Table \ref{tab:result} showcases the effectiveness of our framework. PTM-SIB-KMeans$_{FT}$ exhibited substantial superiority over state-of-the-art models and baselines. With BERT as the core PTM, we observed significant enhancements across DBpedia, AG News, SMS Spam, and ISEAR, with performance gains of 45.87\%, 20.16\%, 12.54\%, and 9.98\%, respectively, compared to PTMs. The results for Yahoo and Polarity remarkably improved when utilizing a RoBERTa-based model. Furthermore, we compared the outcomes when labels were absent, where our framework outperformed all benchmark datasets.
 

In addition to our proposed framework, we also compared our results to those obtained using existing state-of-the-art models, namely BERT$_{IT:CLUSTER}$, BERT$_{IT:MLM}$, and BERT$_{IT:MLM+CLUSTER}$. Our Be-SIB-KMeans$_{FL}$ framework, based on BERT, demonstrated superior performance for DBpedia, ISEAR, and AG News, achieving accuracy gains of at least 8.41\%, 3\%, and 3.34\%, respectively. It also showed slight improvements for SMS spam compared to BERT$_{IT:CLUSTER}$. When using RoBERTa as the base model, the results for Yahoo Answers and Polarity improved by 3\% and 7\%, respectively. Overall, our framework consistently produced more stable results and outperformed the state-of-the-art models using three different PTMs as base models. While BERT$_{IT:MLM+CLUSTER}$ achieved comparable results for ISEAR, Subjectivity, and SMS Spam, it failed to surpass the overall performance of our framework. We attribute these superior results to the transfer of knowledge from PTM-SIB to PTM-SIB-KMeans, which utilizes a small fraction of text, sentence embedding support, and clustering impediments. Additionally, we compared our few-label Be-SIB-KMeans$_{FL}$ results to state-of-the-art few-shot learning models. Notably, our few-label approach exhibited superior performance on Subjectivity and AG News datasets, likely due to the combination of randomization, two-stage clustering, and inter-training techniques.

The PTM-SIB-KMeans$_{FT}$ model, which utilizes a combination of different text embedding methods, demonstrates superior robustness and effectiveness on multi-class classification tasks compared to binary ones. Notably, incorporating only a fraction of the text into the KMeans clustering process yielded enhanced model performance compared to using the entire text.


\begin{table}
\footnotesize
\centering
\caption{Comparing Be-SIB-KMeans$_{FT}$ with the full training set in the second stage of Be-SIB-KMeans and Be-SIB-SIB }
\label{tab:our models}
\begin{tabular}{p{1.8cm} p{1.2cm}p{1.2cm}p{1.2cm} }
\hline \hline 
{  Datasets}      & {  Be-SIB-KMeans}  & {  Be-SIB-SIB} & \multicolumn{1}{p{1.2cm}}{{  Be-SIB-KMeans$_{FT}$}} \\ \hline 
 {  SMS Spam}      & {  97.84}          & {  93.54}      & {  \textbf{98.74}}                       \\
{  Polarity  }    & {  59.13 }         & {  65.41 }     & {  \textbf{69.82}}                       \\
{  Subjectivity}  & {  86.40}           &  {  92.70}       &  {  \textbf{93.80}}                       \\
 {  Yahoo Answers} &  {  \textbf{54.26}} &  {  46.24 }     &  {  52.54 }                               \\
 {  ISEAR }         & {  26.49  }        &  {  31.29}      & {  \textbf{32.88}}                        \\
 {  AG News }       &  {  81.42     }     &  {  81.84  }    &  {  \textbf{84.64} }                      \\
 {  DBpedia }      &  {  76.10} &  {  69.29}      &  {  \textbf{77.41} } \\ 
 \hline \hline               
\end{tabular}
\end{table}

\begin{figure}
\centering
  \includegraphics[scale=0.45]{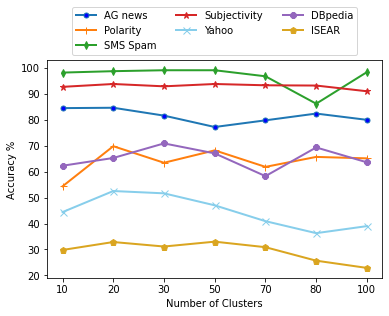}
  \caption{Number of clusters For Be-SIB-KMeans$_{FT}$}
  \label{fig:model_clusters}
\end{figure}

\subsection{Result Analysis}
This section delves into the rationale behind selecting the appropriate number of clusters and elucidates how our model facilitates knowledge transfer by leveraging the clustering algorithm.

We meticulously investigate the optimal number of clusters to be incorporated into our model. The inherent disparity between the number of classes and the size of the datasets being evaluated necessitates the identification of a suitable cluster count. We employ the ideal number of clusters, denoted as $C$, for the clustering algorithms SIB and KMeans, based on the data statistics. Figure \ref{fig:model_clusters} reveals that the performance of the clustering algorithms diminishes when $C$=$\{70,80,100\}$ for the ISEAR and Yahoo Answers datasets. However, employing 20 clusters for SIB and KMeans yielded a significant improvement, registering gains of 32.88\% and 52.54\%, respectively. Interestingly, reducing $C$ to 10 fails to enhance performance for the Polarity, DBpedia, and Yahoo Answers datasets. AG News and Subjectivity exhibit minimal sensitivity to cluster size variations.

Our selection of 20 clusters is guided by the average performance across all datasets for each $C$ This approach yielded accuracy scores of 66.61\%, 71.10\%, 70.11\%, 69.35\%, 65.97\%, 65.55\%, and 65.71\% for cluster counts of 10, 20, 30, 50, 70, 80, and 100, respectively. Consequently, the highest accuracy of 71.10\% is achieved when $C$= 20. This strategy of meticulously setting the cluster hyperparameter promotes stability and balance within our model, a common practice in domain adaptation scenarios.

 \begin{figure*}[!htb]
 \centering
  \includegraphics[scale=0.50]{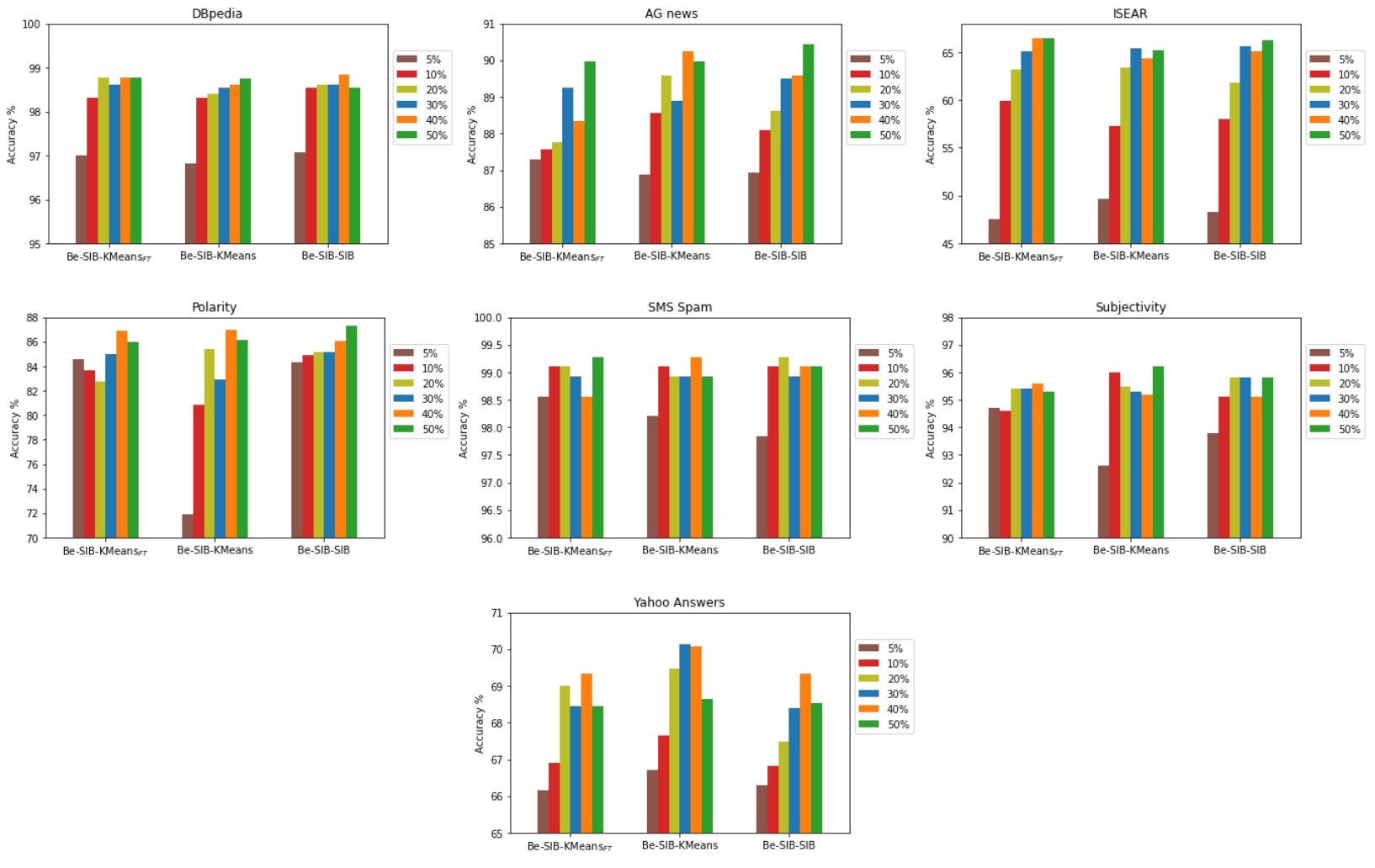}
  \caption{Accuracy results for our model Be-SIB-KMeans$_{FT}$, Be-SIB-KMeans and Be-SIB-SIB with limited samples in the fine-tune model.}  
  \label{fig:result}
\end{figure*}

\begin{table}
\centering
\scriptsize
\caption{Normalized Mutual Information (NMI) for clusters and accuracy for interior models }
\label{tab:nmi-result}
\begin{tabular}{p{1.8cm}p{1.0cm}p{1.0cm}p{1.0cm}p{1.0cm}}
\hline \hline
{\textbf{Dataset} }        & \multicolumn{2}{c}{ { \textbf{NMI}}}              & \multicolumn{2}{c}{{ \textbf{Accuracy (\%)}}}         \\ 

               & \multicolumn{1}{p{1.0cm}}{\emph{SIB}} & \multicolumn{1}{p{1.0cm}}{\emph{KMeans}} & \multicolumn{1}{p{1.0cm}}{{\textbf{Be-SIB}}} & \multicolumn{1}{p{1.0cm}}{{ \textbf{Be-SIB-KMeans}}} \\
{  AG News }      & {  0.399    }       & {  0.3725}             & {  23.09}                     &{  37.42}                           \\
{  ISEAR }         & {  0.105 }          & {  0.177 }             & {  16.77 }                   & {  17.04  }                         \\
{  Polarity}       & {  0.0169   }       & {  0.0232 }            &{  40.67}                    & {  48.54 }                          \\
{  SMS Spam }      & {  0.1248 }         & {  0.1089}             & {  94.44 }                  & {  85.84}                           \\
{  Subjectivity   } & {  0.1323 }          &{  0.1053 }            & {  68.40 }                   & {  57.50  }                         \\
{ Yahoo Answers} & {  0.3485}         &{  0.3911  }           &{  5.50  }                   & {  11.85  }                         \\
{  DBpedia  }       & {  0.7573   }       & {  0.7280 }            &{   9.30   }                 & 
 {  14.00} \\
\hline \hline 
\end{tabular}
\end{table}

\begin{table} 
\scriptsize
\centering
\caption{\emph{PTM-SIB-KMeans$_{FT}$} with various fractions of text used in the second stage \emph{PTM-SIB-KMeans}}
\label{tbl:bert-sib-comp}
\begin{tabular}{p{2cm}p{1.2cm}p{1.2cm}p{1.2cm}}
\hline  \hline
\multicolumn{1}{c}{\multirow{2}{*}{\textbf{Datasets}}} & \multicolumn{3}{c}{ \textbf{Be-SIB-KMeans$_{FT}$}}                               \\
\multicolumn{1}{c}{}                                   & \textit{ \textbf{5 \%}} & \textit{ \textbf{10\%}} & \textit{ \textbf{20\%}} \\
\midrule 
{ AG News}                                                & {\  84.64}                  & {  81.42}                  & {  84.60}                  \\
{  ISEAR   }                                               & {  32.88}                  &{  26.49 }                 & {  32.49}                  \\
{  Polarity }                                              & {  69.82}                  &{  59.13}                  & {  70.66 }                 \\
{  SMS Spam }                                              & {  98.74 }                 & {  97.84}                  &{  97.84 }                 \\
{  Subjectivity }                                          & {  93.8}                   & {  86.40}                  & {  92.20}                  \\
{  Yahoo Answers}                                          & {  52.54 }                 & {  54.26 }                 & {  48.34}                  \\
{  DBpedia }                                               & {  77.41}                  & {  75.96 }                 & {  64.11}    \\

\multicolumn{1}{c}{\multirow{2}{*}{\textbf{}}} & \multicolumn{3}{c}{ \textbf{RoB-SIB-KMeans$_{FT}$}}                              \\
\multicolumn{1}{c}{}                                   & \textit{ \textbf{5 \%}} & \textit{ \textbf{10\%}} & \textit{ \textbf{20\%}} \\
\midrule 
{  AG News}                                                & {  82.64}                  &{  85.60}                  & {  85.06}                  \\
{  ISEAR}                                                  & {  32.62}                  & {  35.28 }                 & {  37.68}                  \\
{  Polarity }                                              & {  75.82}                  & {  71.13  }                & {  62.79 }                \\
{  SMS Spam}                                               &{  98.20 }                 & {  98.74}                  &{  98.38 }                \\
{  Subjectivity }                                          & {  92.00}                   & {  90.80}                  & {  92.00}                  \\
{  Yahoo Answers }                                         & {  54.54}                  & {  59.26}                  & {  62.76 }                \\
{  DBpedia}                                                & {  74.42}                  & {  78.48}                  & {  84.83 }   \\
\multicolumn{1}{c}{\multirow{2}{*}{\textbf{}}} & \multicolumn{3}{c}{ \textbf{DisB-SIB-KMeans$_{FT}$}}                               \\
\multicolumn{1}{c}{}                                   & \textit{ \textbf{5 \%}} & \textit{ \textbf{10\%}} & \textit{ \textbf{20\%}} \\
\midrule 
{  AG News}                                               & {  78.86}                  &{  78.16}                  & {  79.42 }                 \\
{  ISEAR  }                                                & {  27.29}                  & {  28.36 }                 & {  27.56}                  \\
{  Polarity }                                              & {  69.82}                  &{  59.13  }                & {  70.66 }                 \\
{  SMS Spam }                                              &  {  96.65 }                 & {  97.67 }                 & {  97.49 }                 \\
{   Subjectivity}                                           &{   91.60 }                 & {   90.50 }                 & {   91.60}                  \\
{   Yahoo Answers }                                          & {  40.83 }                 & {   44.48 }                & {   49.65}                  \\
{  DBpedia }                                               & {  73.02}                  & {   74.14 }                 & {   78.90}    \\
\hline \hline
\end{tabular}

\centering
\caption{\emph{PTM-SIB-KMeans$_{FT}$} with various fractions of text used in the first stage \emph{PTM-SIB}}
\label{tab:differnt_input}
\begin{tabular}{p{2cm}p{1cm}p{1cm}p{1cm}p{1cm}}
\hline  \hline
\multicolumn{1}{c}{\multirow{2}{*}{\textbf{Datasets}}} & \multicolumn{4}{c}{\textbf{Be-SIB-KMeans$_{FT}$}}                                                         \\ 
\multicolumn{1}{c}{}                                   & \textit{\textbf{100\%}} & \textit{\textbf{80\%}} & \textit{\textbf{70\%}} & \textit{\textbf{50\%}} \\
\midrule 
AG News                                                & 84.64                   & 84.78                  & 82.78                  & 80.54                  \\
ISEAR                                                  & 32.88                   & 27.29                  & 27.16                  & 30.49                  \\
Polarity       the                                         & 69.82                   & 61.29                  & 66.35                  & 61.85                  \\
SMS Spam                                               & 98.74                   & 94.98                  & 98.56                  & 99.10                   \\
Subjectivity                                           & 93.82                    & 92.46                   & 93.91                   & 91.60                   \\
Yahoo Answers                                          & 52.54                   & 47.53                   & 46.10                   & 48.62                  \\
DBpedia                                                & 77.41                   & 68.31                  & 70.18                  & 67.80  \\ 

\multicolumn{1}{c}{\multirow{2}{*}{\textbf{}}} & \multicolumn{4}{c}{\textbf{RoB-SIB-KMeans$_{FT}$}}                                                         \\ 
\multicolumn{1}{c}{}                                   & \textit{\textbf{100\%}} & \textit{\textbf{80\%}} & \textit{\textbf{70\%}} & \textit{\textbf{50\%}} \\
\midrule 
AG News                                                & 82.64                   & 82.54                  & 84.36                  & 79.51                  \\
ISEAR                                                  & 32.62                   & 33.82                  & 35.28                  & 32.62                 \\
Polarity                                          & 75.82                  & 72.35                  & 74.78                  & 65.97                  \\
SMS Spam                                               & 98.20                   & 98.38                  & 98.30                  & 98.02                   \\
Subjectivity                                           & 92.00                    & 91.40                   & 89.80                   & 92.91                   \\
Yahoo Answers                                          & 54.54                   & 54.59                   & 57.20                   & 54.00                  \\
DBpedia                                                & 74.42                  & 77.32                  & 79.56                 & 76.15  \\ 

\multicolumn{1}{c}{\multirow{2}{*}{\textbf{}}} & \multicolumn{4}{c}{\textbf{DisB-SIB-KMeans$_{FT}$}}                                                         \\ 
\multicolumn{1}{c}{}                                   & \textit{\textbf{100\%}} & \textit{\textbf{80\%}} & \textit{\textbf{70\%}} & \textit{\textbf{50\%}} \\
\midrule 
AG News                                                & 78.86                   & 83.90                  & 78.02                  & 78.90                  \\
ISEAR                                                  & 27.29                   & 27.16                  & 27.96                  & 26.09                  \\
Polarity                                                & 63.26                   & 63.16                  & 63.82                  & 62.41                  \\
SMS Spam                                               & 96.95                   & 96.77                  & 96.59                 & 95.87                   \\
Subjectivity                                           & 91.60                    & 91.7                   & 92.00                   & 90.80                   \\
Yahoo Answers                                          & 40.83                   & 45.87                   & 47.87                   & 47.92                  \\
DBpedia                                                & 73.02                   & 70.97                  & 63.69                  & 71.86  \\ 
\hline \hline
\end{tabular}
\end{table}

We propose the potential of knowledge transfer within unsupervised methods by leveraging text fractions within the intermediate schema when there is a restricted number of labels. Initially, we quantify the quality of the individual models (PTM-SIB and PTM-SIB-KMeans) by adopting Normalized Mutual Information (NMI). NMI quantifies the amount of transferable knowledge between clusters. NMI operates on the normalized level and is calculated as follows:

\begin{displaymath}
  NMI(P_i,P_j) = \frac{I(P_i,P_j)}{\sqrt{H(P_i)H(P_j)}}
\end{displaymath}

\noindent, where $P_i$ is a true label and $P_j$, is a pseudo-label for SIB and KMeans. Mutual information between $P_i$ and $P_j$ is defined by $I(.)$ and the entropy for $P_i$ and $P_j$ is denoted by $H(P_i)$ and $H(P_j)$, respectively. Table \ref{tab:nmi-result} shows that KMeans achieves similar NMI scores to SIB where only a small portion of the input text is used and indicates convergence quality.

To assess the knowledge transformation achieved through correction, we treated both Be-SIB and Be-SIB-KMeans as unsupervised pre-trained models with the BERT-base model and evaluated their performance on a test set. Table \ref{tab:nmi-result} summarizes the accuracy results, demonstrating the learning benefits within the internal model. Despite the limited input text for KMeans, the results indicate that NMI scores are either comparable to or superior to those obtained with the SIB algorithm. This suggests that the second stage of the unsupervised pre-trained model effectively captures the acquired knowledge, as evident by its improved performance. For instance, Be-SIB-KMeans achieved a remarkable accuracy boost of 14.33\% and 6.35\% for the AG News and Yahoo Answers datasets, respectively. However, it encountered lower accuracy for SMS Spam and Subjectivity, likely due to the restricted text size and the relatively small number of classes.

\subsection{Ablation Study}
To further investigate the effectiveness of our model, we conducted ablation studies to identify the key components that contribute to its performance.

Table \ref{tab:our models} summarizes the results obtained by duplicating the same cluster Be-SIB-SIB with full-text input for both clusters and utilizing a different cluster for Be-SIB-KMeans. The hyperparameters $T_n$ and $T_s$ were set to 100\%, and the fine-tuned model parameters $T_m$ and $P_m$ were set to 64 samples. Overall, our model achieved superior performance across most datasets. Be-SIB-KMeans$_{FL}$ consistently outperformed the other two variants. On the Polarity, Subjectivity, ISEAR, and AG News datasets, our model demonstrated significant improvements of 10.70\%, 7.40\%, 6.39\%, and 3.22\%, respectively. SMS Spam saw a marginal gain of less than 1\% in accuracy. However, the intricate text structure of Yahoo Answers necessitated full-text training. Replicating cluster algorithms like SIB with TF-IDF resulted in performance degradation for SMS Spam, Yahoo Answers, and DBpedia.


Additionally, we conducted experiments with varying percentages of labelled data to assess the impact of label quantity on our model's performance. As shown in Figure \ref{fig:result}, our model consistently exhibited superior performance until reaching 20\% of labelled examples. Notably, our model demonstrated the highest degree of stability across the ISEA and DBpedia databases compared to the other models. The Be-SIB-KMeans$_{FT}$ model for AG News achieved similar results for 5\%, 10\%, and 20\% due to its limited training on a small subset during the second stage. When 50\% of the labelled data is released, Be-SIB-KMeans and Be-SIB-SIB attained comparable results to nBe-SIB-KMeans$_{FT}$ for most datasets. However, binary class datasets exhibited performance degradation when the sample size exceeded 5\%. The shared contextual meaning among classes in the Polarity and Subjectivity datasets confounded our model, as did the number of clusters. This suggests that our model is better suited for multi-class tasks and less suitable for binary class scenarios.


\begin{figure}[htbp]
\centering
  \includegraphics[scale=0.40]{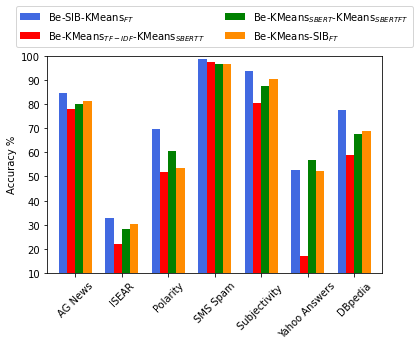}
  \caption{Different components IDoFew are presented in the ablation study.}
  \label{fig:Ablation}
\end{figure}

The sequence of the cluster algorithms is taken into consideration in our proposed framework as shown in Figure \ref{fig:Ablation}. The parameters of all models are set as fixed for a fair evaluation, namely the full text is used for the first clusters, 5\% is used for the second, and 64 samples are used for the fine-tuning stage. For Be-KMeans-SIB$_{FT}$, we replace SIB with KMeans along with SBERT embedding followed by SIB. The performance is asymptotic in result for most datasets to the Be-SIB-KMeans$_{FT}$. The previous models' prime goal is to show the robustness of KMeans and SIB. We repeated KMeans with the same and different embedding techniques. However, the performance of Be-KMeans-KMeans$_{FT}$ drops for all datasets compared with Be-SIB-KMeans$_{FT}$. Overall, it is not effective that repeating the same cluster algorithm will give us the same performance even though diverse embedding techniques are utilized.

The text input of the training set plays an important role in our model which leads us to conduct further investigations for both cluster stages SIB and KMeans which are based on PTMs. Table \ref{tbl:bert-sib-comp} shows the fraction size of the text used in the second stage. The results show that the number of classes might lead to the choice of a better fraction of text. For example, fewer classes such as SMS Spam, Subjectivity, and Polarity create uncertainty for our model. Multi-classes perform much better when more text is released. Table \ref{tab:differnt_input} presents the fraction of the text used in the first cluster. As expected, most of the datasets along with PTMs are showing a drop in results. More text is needed in the first stage to create the base knowledge.  

\section{Conclusions and Limitations}
We present a novel dual-clustering algorithm, IDoFew, that interweaves label-efficient learning with PTMs to enhance text classification performance. Our approach effectively leverages the complementary strengths of two clustering models while minimizing computational overhead. We showcase the robustness of using an unsupervised PTM model within an intermediate stage to facilitate knowledge transfer throughout the clustering process. Remarkably, our algorithm achieves superior performance using only a small fraction of text (5\%) for training, demonstrating the power of knowledge transfer with PTMs. Despite varying text structures and class distributions across datasets, our proposed PTMs-SIB-KMeans$_{FT}$ variant consistently outperforms state-of-the-art models.

One potential improvement lies in fine-tuning the number of clusters to further enhance results. Additionally, exploring semantic clustering engines like probabilistic topic models, such as Latent Dirichlet Allocation (LDA) \cite{blei2003latent}, could further boost performance, although it might introduce some computational overhead.

\begin{acks}
This work is partially supported by the Australian Research Council (ARC) under Grant number DP220103717 and the National Science Foundation of China under Grant number 62072257.
\end{acks}

\bibliographystyle{ACM-Reference-Format}
\balance
\bibliography{references}
\end{document}